\pdfoutput=1

\documentclass[11pt]{article}

\usepackage[]{acl}

\usepackage{times}
\usepackage{latexsym}

\usepackage[T1]{fontenc}

\usepackage[utf8]{inputenc}

\usepackage{microtype}

\usepackage{graphicx}
\usepackage{amsmath}
\usepackage{hyperref}
\usepackage{bbm}
\usepackage{CJKutf8}
\usepackage{cleveref}
\usepackage{graphicx}
\usepackage{url}

\newcommand{\pararel}{\textsc{ParaRel}\raisebox{-2pt}{\includegraphics[width=0.15in]{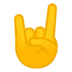}}}
\newcommand{\mpararel}{\textsc{mParaRel}}
\DeclareMathOperator*{\argmax}{arg\,max}
\crefformat{section}{\S#2#1#3}
\crefformat{subsection}{\S#2#1#3}
\crefformat{subsubsection}{\S#2#1#3}

\title{Factual Consistency of Multilingual Pretrained Language Models}

\author{First Author \\
  Affiliation / Address line 1 \\
  Affiliation / Address line 2 \\
  Affiliation / Address line 3 \\
  \texttt{email@domain} \\\And
  Second Author \\
  Affiliation / Address line 1 \\
  Affiliation / Address line 2 \\
  Affiliation / Address line 3 \\
  \texttt{email@domain} \\}
  
\author{Constanza Fierro ~ Anders Søgaard \\
  University of Copenhagen \\
  \texttt{\{c.fierro,soegaard\}@di.ku.dk}, \\
 }

\begin{document}
\maketitle
\begin{abstract}
Pretrained language models can be queried for factual knowledge, with potential applications in knowledge base acquisition and tasks that require inference. However, for that, we need to know how reliable this knowledge is, and recent work has shown that monolingual English language models lack {\em consistency} when predicting factual knowledge, that is, they fill-in-the-blank differently for paraphrases describing the same fact. In this paper, we extend the analysis of consistency to a {\em multilingual} setting. We introduce a resource, \mpararel\footnote{\url{https://github.com/coastalcph/mpararel}}, and investigate (i) whether multilingual language models such as mBERT and XLM-R are more consistent than their monolingual counterparts;
and (ii) if such models are equally consistent across languages.
We find that mBERT is as inconsistent as English BERT in English paraphrases, but that both mBERT and XLM-R exhibit a high degree of inconsistency in English and even more so for all the other 45 languages.
\end{abstract}

\section{Introduction}

Pretrained Language Models (PLMs)
enable high-quality sentence and document representations~\citep{peters-etal-2018-deep, devlin-etal-2019-bert, yang2019xlnet, JMLR:v21:20-074} 
and encode world knowledge that can be useful for downstream tasks, e.g. closed-book QA~\citep{roberts-etal-2020-much}, and commonsense reasoning~\citep{zellers2019recognition, talmor-etal-2019-commonsenseqa}, to name a few. Recent work has used language models as knowledge bases
~\citep{petroni-etal-2019-language, kassner-etal-2021-multilingual, roberts-etal-2020-much} and as the basis of neural databases~\citep{neural-databases-thorne}. 
Such usage of PLMs relies on the assumption that we can generally trust the world knowledge that is induced from these models.
Consistency is a core quality that we would like models to have when we use their stored factual knowledge. We want models to behave consistently on semantically equivalent inputs~\citep{Elazar2021MeasuringAI}, and to be consistent in their believes~\citep{kassner-etal-2021-beliefbank}. Moreover we want them to be fair across languages or in other words to exhibit a consistent behaviour across languages~\citep{Choudhury_Deshpande_2021}.
Nonetheless, recent work on consistency in PLMs has shown that models are brittle in their predictions when faced to irrelevant changes in the input~\citep{gan-ng-2019-improving,ribeiro-etal-2020-beyond,Elazar2021MeasuringAI,ravichander-etal-2020-systematicity}. These works only considered English PLMs, while \citet{jang2021accurate} studied the consistency of Korean PLMs. There are, to the best of our knowledge, no resources available to measure the consistency of multilingual PLMs. 

\paragraph{Contributions} In this paper, we present \mpararel{}, a multilingual version of the \pararel{} dataset~\citep{Elazar2021MeasuringAI}, which we construct by automatically translating the English data to 45 languages and performing a human review of 11 of these. We then evaluate how consistent mBERT is in comparison to its monolingual counterpart, and we study how the consistency of mBERT and XLM-R varies across different languages. Following previous work, we do this by querying the model with cloze-style paraphrases, e.g. ``Albert Einstein was born in [MASK]'' and ``Albert Einstein is originally from [MASK]''. We find that mBERT and XLM-R exhibit competitive consistency to English BERT, but consistency numbers are considerably lower for other languages. In other words, while consistency is a serious problem in PLMs for English \citep{Elazar2021MeasuringAI}, it is a {\em much} bigger problem for other languages. 

\section{Probing Consistency}
We use the same probing framework as defined by \citet{petroni-etal-2019-language} and refined by \citet{Elazar2021MeasuringAI}, and query PLMs with cloze-test statements created from subject-relation-object Wikidata triples~\citep{elsahar-etal-2018-rex}. That is, we have a set of different relations $\{r\}$, and each $r$ has a set of templates or patterns $\{t\}$ and a set of subject-object tuples $\{(s,o)\}$. Each template $t$ describes its corresponding relation $r$ between the pairs $(s,o)$. E.g. a relation $r$ can be \textit{born-in}, and two patterns could be $\{t_1=$``[X] was born in [Y]'', $t_2=$``[X] is originally from [Y]''$\}$ (where [X] is the subject and [Y] the object to be replaced). Then the corresponding subject-object tuples $\{(s,o)\}$ are used to query and evaluate the model by replacing the subject and masking the object. We study the consistency of a PLM by querying it with cloze-test paraphrases and measuring how many of the predictions of the paraphrases are the same (details in \cref{sec:evaluation_metrics}).

\section{\mpararel{}}
We used the paraphrases in the \pararel{} dataset~\citep{Elazar2021MeasuringAI}, which has 38 relations in total and an average of 8.6 English templates per relation. We translated these using the procedure below, obtaining paraphrases for 46 languages.

\paragraph{Translations}
We relied on five different machine translation models: Google Translate\footnote{\url{https://pypi.org/project/googletrans/}}, Microsoft Translator\footnote{\url{https://docs.microsoft.com/en-us/azure/cognitive-services/translator/}}, a pretrained mBART model that translates between 50 languages~\citep{tang2020multilingual}, a pretrained mixture of Transformers that translates between 100 languages~\citep{fan2021beyond}, and  OPUS-MT~\citep{tiedemann-thottingal-2020-opus}. We fed models with templates, e.g.,``[X] died in [Y]''\footnote{Translating populated templates made alignment hard.}, automatically checking if the translation contained [X] and [Y]. We considered as valid: (1) translated paraphrases that were agreed upon by two or more different models, and (2) the translations from the Microsoft translator, as they were found to be of good quality in several languages as per manual inspection by native speakers. So for languages that Microsoft supports, we will have a template $t$ from the Microsoft translator, as well as any other translation agreed upon by two or more other translators\footnote{In the final dataset, 60\% of the templates are agreed by 2 or more translators}. Finally, we also include the templates in the mLAMA dataset~\citep{kassner-etal-2021-multilingual}. Translations of subject-object entities were obtained from WikiData, using the entity identifiers. We kept only the languages that (i) covered at least 60\% of the total 38 relations,\footnote{Only relations with more than one template with subject-object tuples were included.}, and (ii) covered at least 20\% of the total original phrases in English.\footnote{A phrase is a populated template.}

\begin{table}[]
\centering
\scalebox{0.85}{
\begin{tabular}{lr}
\hline
\textbf{}                      & \textbf{\mpararel{}} \\ \hline
Average \#relations                           & 37.13 \\
Average total \#patterns                      & 343   \\
Min. patterns in a relation                   & 2     \\
Max. patterns in a relation                   & 33    \\
Average patterns in a relation                & 9.2   \\
Average string distance                       & 13.9  \\ \hline
\end{tabular}
}
\caption{\mpararel{} statistics across languages.}
\label{tab:mpararel_stats}
\end{table}

\begin{figure}[]
\centering
\includegraphics[width=0.48\textwidth]{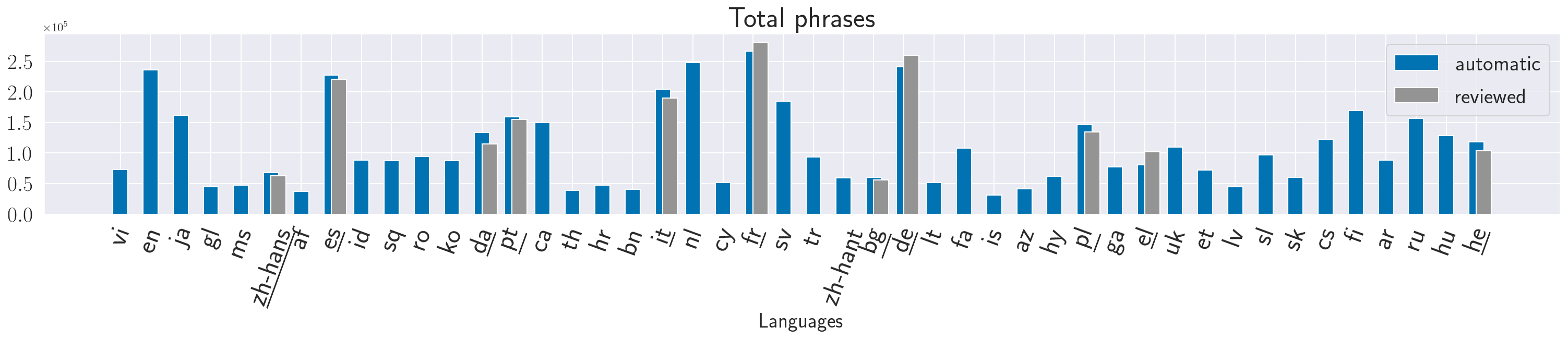}
\caption{Number of examples per language. Manually reviewed languages are underlined. The order is given by the consistency results (see Figure~\ref{fig:mpararel_results}).}
\label{fig:total_phrases}
\end{figure}

\paragraph{Human Evaluation}\label{sec:human_evaluation}
For assessing the quality of the translated paraphrases we carried out a human review. We had 14 native speakers review 11 different languages\footnote{There were 2 reviewers in Greek, German, and Spanish.}. Each person reviewed a 50\% random sample of the total templates of the language\footnote{The review took 50 minutes on average and the reviewers did it voluntarily.}. We asked whether the template was a correct paraphrase of the given relation, we requested corrections and optionally asked for new template suggestions. On average, 16\%$\pm$8\% of the reviewed templates were considered wrong, 20\%$\pm$10\% were amended, and the rest were considered correct. The statistics of the dataset after removing the wrong templates and including the corrections and suggestions can be found in Table~\ref{tab:mpararel_stats}. The total number of different phrases (templates with the subject and object replaced) per language is shown in Figure~\ref{fig:total_phrases}.

\section{Experiments}
We ran experiments with mBERT~\citep{devlin-etal-2019-bert}, a multilingual BERT model of 110M parameters trained on 104 languages using Wikipidea, and XLM-RoBERTa~\citep{conneau-etal-2020-unsupervised}, a multilingual RoBERTa model of 560M parameters trained on 100 languages using 2.5TB of CommonCrawl data.

\paragraph{Querying Language Models}\label{sec:query_lm}
The prediction of a PLM for a cloze statement $t$ is normally $\argmax_{w \in V}(w|t)$~\citep{petroni-etal-2019-language, ravichander-etal-2020-systematicity}, that is, the top-1 token prediction over the vocabulary. However, \citet{kassner-etal-2021-multilingual,Elazar2021MeasuringAI} used typed queries, where the prediction is $\argmax_{w \in C}(w|t)$, with $C$ a set of candidates that meets the type criteria of the pattern (e.g. cities, professions). In our case,
$C$ is all the possible objects in the relation. The motivation is that by restricting the output we can reduce the errors due to surface fluency, as when populating the template with different tuples small grammatical errors can occur~\citep{kassner-etal-2021-multilingual}.

It is common to only consider tuples (subject-object) for which the to-be-masked object is a single token in the models vocabulary~\citep{petroni-etal-2019-language,Elazar2021MeasuringAI}. However, this reduces the number of valid tuples severely, and even more so when dealing with multilingual vocabularies. Therefore, we follow the multi-token prediction approach in \citet{kassner-etal-2021-multilingual} and query the model with multiple masked tokens. The probability of an object instantiation is then the average probability of its tokens, i.e., for a given object $o=w_1w_2...w_l$,
$p(o|t) = \frac{1}{l}\sum_{i=1}^l p(m_i=w_i|t_l)$, where $w_i$ is the $i$-th token of the word $o$,  $m_i$ is the $i$-th mask token, and $t_l$ is the template with $l$ mask tokens.

\paragraph{Evaluation}\label{sec:evaluation_metrics}
For a given relation $r$ the {\bf consistency} is the percentage of pairs of templates that have the same prediction for every subject-object tuple~\citep{Elazar2021MeasuringAI}, i.e. the consistency of a given relation $r$ is:
\begin{align}\label{math:consistency}
    \frac{1}{|D|}\sum_{d\in D}\frac{2}{|T|(|T|-1)}\sum_{i=0}^{|T|} \sum_{j=i+1}^{|T|}\mathbbm{1}_{f(t_i^d) = f(t_j^d)}
\end{align}
where $t$ is a template, $T$ the set of templates in the relation, $d$ is a subject-object tuple, $D$ the set of all tuples, so $t_i^d$ is the $i$-th template populated with the subject-object data $d$, and $f(\cdot)$ is the prediction of the model. Next, {\bf accuracy} measures the factual correctness of the predictions and is defined as the percentage of correct predictions over all the templates and data, i.e. $\sum_{d \in D}\sum_{t \in T}\mathbbm{1}_{f(t^d) = o}$, where $o$ is the object of the tuple $d$. Finally, \textbf{consistency-accuracy} is the subset of the accurate predictions that is also consistent. Thus, it is computed similarly to Equation~\ref{math:consistency} but in the indicator's condition we also add the condition imposed in the accuracy. This metric is useful to account for trivial cases of consistency: A model can be really bad in a language and predict the same token despite the input, and thus be perfectly consistent. For all metrics, we report the {\em macro average} across relations.\footnote{Our results are {\em not} directly comparable to those reported in  \citet{Elazar2021MeasuringAI}, even if we use the same metric, since we filter tuples with the same subject, but two different objects.}
\begin{table}[]
\centering
\scalebox{0.8}{
\begin{tabular}{llcccc}
\hline
Metric          &       & BERT          & \multicolumn{3}{c}{mBERT}                     \\
                &       & en            & en            & ja            & zh-hans       \\ \hline
Consistency     & w/ .  & \textbf{0.57} & \textbf{0.54} & \textbf{0.55} & 0.46          \\
                & w/o . & 0.53          & 0.53          & 0.52          & \textbf{0.51} \\
Accuracy        & w/ .  & \textbf{0.39} & \textbf{0.37} & 0.13          & 0.22          \\
                & w/o . & 0.32          & 0.35          & \textbf{0.15} & \textbf{0.27} \\
Consistency-acc & w/ .  & \textbf{0.32} & \textbf{0.3}  & 0.09          & 0.15          \\
                & w/o . & 0.24          & 0.28          & \textbf{0.1}  & \textbf{0.2}  \\ \hline
\end{tabular}
}
\caption{Performance comparison of BERT to mBERT, as well as of removing sentence-final punctuation in our input examples, with mBERT results on English, Japanese, and Chinese Simplified. }
\label{tab:point_vs_no_point}
\end{table}

\begin{figure*}[]
\centering
\includegraphics[width=1.\linewidth]{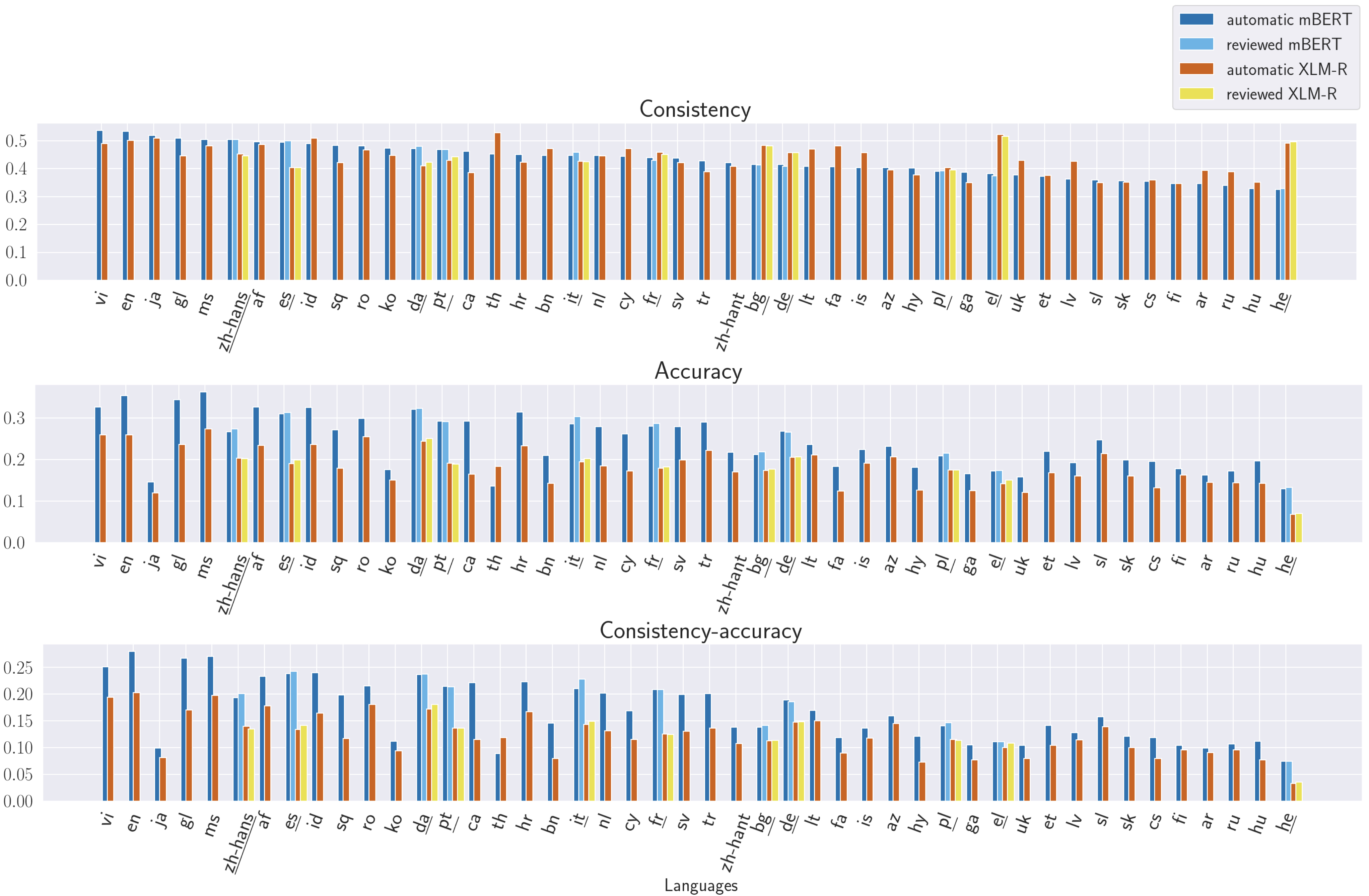}
\caption{mBERT and XLM-R results on \mpararel{} after and before the human review (\cref{sec:human_evaluation}). The order of the languages follows the consistency results in mBERT, and the languages underlined were manually reviewed.}
\label{fig:mpararel_results}
\end{figure*}

\section{Results and Discussion}

Table~\ref{tab:point_vs_no_point} compares the consistency of BERT and mBERT on English data, showing little to no difference, depending on whether we use sentence-final punctuation or not. Sentence-final punctuation is not fully consistent in the machine translation output, so we ran experiments comparing the performance of including sentence-final punctuation or removing it. Since languages vary in how they use punctuation, and sentence-final punctuation causes variance in consistency (e.g., Japanese +3\%, but Chinese Simplified -5\%), we decided to remove all sentence-final punctuation for the cross-lingual consistency results. 

\paragraph{Consistency across languages}\label{sec:consistency_results}
The consistency results in the \mpararel{} dataset are presented in Figure~\ref{fig:mpararel_results}. First of all, we can see that the manual corrections don't change the results much (as also experienced by~\citet{kassner-etal-2021-multilingual}). Nevertheless, they do improve the consistency and accuracy by 1\%-2\% in a couple of languages, probably because some noise was reduced when correcting and adding new templates.
Consistency numbers remain very low, however, especially for other languages than English and Vietnamese.
XLM-R is much more consistent than mBERT in {\em some} languages (e.g. Greek (`el')), yet their average consistency is the same (0.43). The standard deviation of XLM-R's consistency is 8\% lower than that of mBERT, i.e., XLM-R has a more fairly distributed consistency.
Somewhat surprisingly, the accuracy of mBERT is superior to XLM-R's, nevertheless, this aligns to the findings of~\citet{Elazar2021MeasuringAI} where English base BERT obtained higher accuracy than a large English RoBERTa model. We note the importance of controlling for accuracy in our consistency results (reported as {\em consistency-accuracy}): Japanese, for example, has high consistency, but in part, because it wrongly predicts the same (frequent) token across paraphrases; consistency-accuracy reranks Japanese as one of the most inconsistently encoded languages in both mBERT and XLM-R. 

\section{Related Work}
\citet{petroni-etal-2019-language,davison-etal-2019-commonsense} first studied to what extent PLMs store factual and commonsense knowledge, proposing the LAMA probe and dataset. Then further analysis followed it, \citet{kassner-schutze-2020-negated} studied probing PLMs factual knowledge on negated sentences, \citet{shin-etal-2020-autoprompt, reynolds_prompt_programming,jiang-etal-2020-know} optimized the prompts so to improve the knowledge retrieval, and \citet{Bouraoui_Camacho-Collados_Schockaert_2020,heinzerling-inui-2021-language} explored other approaches different than the cloze-test probing. Then, \citet{kassner-etal-2021-multilingual,jiang-etal-2020-x} studied the knowledge memorized in multilingual PLMs, presenting the mLAMA dataset which is a translated version of LAMA.

Consistency in PLMs has been studied in English. \citet{gan-ng-2019-improving} created a paraphrased version of SQuAD and showed that the state-of-the-art models had a significant decrease in performance,~\citet{ribeiro-etal-2020-beyond} proposed a framework to test the robustness in the predictions when faced with irrelevant changes in the input.~\citet{Elazar2021MeasuringAI,ravichander-etal-2020-systematicity} showed that monolingual English PLMs are inconsistent in fill-in-the-blank phrases. Then, \citet{newman2021p} proposed using adapters to better handle this inconsistency.

There are paraphrase datasets available in English~\citep{dolan-brockett-2005-automatically,QQP} and in multiple languages~\citep{Ganitkevitch-Callison-Burch-2014:LREC}, but they cannot be easily linked to subject-object tuples in order to measure consistency.

\section{Conclusion}
In this work, we measured the consistency of \textit{multilingual} Pretrained Language Models when queried to extract factual knowledge. We constructed a high-quality multilingual dataset containing 46 different languages, to assess the consistency of models predictions in the face of language variability. Finally, we experimented with mBERT and XLM-R and concluded that their consistency is poor in English, but even worse in other languages.

\section*{Acknowledgements}
We thank Laura Cabello, Stephanie Brandl, Yova Kementchedjhieva, Daniel Hershcovich, Emanuele Bugliarello, Katerina Margatina, Karolina Stanczak, Ilias Chalkidis, Ruixiang Cui, Rita Ramos, Stella Frank, and Stephanie Brandl for their time and effort invested in reviewing the translations, and the members of the CoAStaL NLP group and the anonymous reviewers for their helpful suggestions.

\bibliography{anthology,custom}
\bibliographystyle{acl_natbib}

\end{document}